\documentclass{article}

\usepackage{interspeech2007,amssymb,amsmath,epsfig}
\usepackage[T1]{fontenc}
\usepackage[T1]{tipa}
\usepackage[latin1]{inputenc}

\ninept
\def\reg{{\rm\ooalign{\hfil
     \raise.07ex\hbox{\footnotesize R}\hfil\crcr\mathhexbox20D}}}

\title{Combined Acoustic and Pronunciation Modelling for Non-Native Speech Recognition}

\name{{\em G. Bouselmi, D. Fohr, I. Illina}}

\address{Speech Group, LORIA-CNRS \& INRIA, ``http://parole.loria.fr/''\\
BP 239, 54600 Vandoeuvre-les-Nancy, France\\
{\small \tt \{ bousselm, fohr, illina \}@loria.fr}}

\begin{document}
\maketitle

\begin{abstract}
In this paper, we present several adaptation methods for non-native speech recognition.
We have tested pronunciation modelling, MLLR and MAP non-native pronunciation adaptation and HMM models retraining on the \textit{HIWIRE} foreign accented English speech database.
The ``phonetic confusion'' scheme we have developed consists in associating to each spoken phone several sequences of confused phones.
In our experiments, we have used different combinations of acoustic models representing the canonical and the foreign pronunciations: spoken and native models,
models adapted to the non-native accent with MAP and MLLR.
The joint use of pronunciation modelling and acoustic adaptation led to further improvements in recognition accuracy.
The best combination of the above mentioned techniques resulted in a relative word error reduction ranging from 46\% to 71\%.


\end{abstract}

\section{Introduction}

Automatic speech recognition (ASR) systems become widely used as their performance constantly increases.
Nevertheless, ASR systems perform poorly when confronted to non-native speakers \cite{ES00,ES4}.
That is, ASR systems are designed to process a spoken language (SL) and their performance drops with non-native speakers, i.e. speakers whose native language (NL) is different from the language they are speaking (SL).
This is due to the fact that these systems are generally not intended to process non-native speech and the databases used in their training do not include foreign accents.

For public services based on ASR, as well as for applications that specifically involve non-native speakers, it is necessary to take into account foreign accents and pronunciation errors.
This issue has been addressed in the literature and several methods have been developed in order to enhance ASR performance with non-native speech.
These methods are based on acoustic or pronunciation modelling, and they vary according to the modifications made to the ASR system.

The acoustic modelling consists in adapting pre-trained acoustic models to better represent non-native accents.
Classical approaches such as MLLR, MAP, and model retraining have yielded some improvements and more sophisticated methods, such as acoustic and pronunciation modelling, allowed further enhancements.
In \cite{ES4}, the authors use a non-native speech database to adapt pre-trained acoustic models of NL.
A manual mapping between SL and NL phones is used in order to translate the canonical transcriptions of an NL-accented SL speech database (i.e. SL speech uttered by speakers having NL as mother tongue).
That is, according to the mapping, each SL phone is replaced by the corresponding NL one.
Then, using those translated transcriptions, NL models are adapted through MLLR, MAP and Baum-welch training on the above mentioned speech database.
Afterwards, the acoustic models of SL are merged with adapted NL models according to an automatically extracted ``phonetic confusion'' matrix.
A ``phonetic confusion'' matrix $M$ is a matrix holding confusion probabilities between two sets of phones : $M(i,j) = P(p_j | p_i)$ is the probability of recognizing a phone $p_j$ from the second set when the phone $p_i$ from the first set was uttered.
On the other hand, the method described in \cite{ESYoon} performs a non-native adaptation in the training process of acoustic models.
For that matter, the authors utilize a standard SL ASR system to establish an intra language ``phonetic confusion'' matrix between SL phones.
This matrix is then employed to tie the triphone models of the confused phones during their training.

Pronunciation modelling, consists in detecting and taking into account non-native pronunciation variants using either phoneticians' knowledge \cite{ES1} or a data-driven procedure \cite{ES2}.
This modelling is then used to modify the lexicon in order to include the non-native pronunciation variants.
In a more recent work \cite{ESJouvet}, the authors use SL phones which were adapted on NL speech and NL phones as pronunciation variants of SL phones.
Furthermore, the authors allow the open and closed, nasal and non-nasal, front and back rounded versions in the pronunciation of vowels.
The lexicon of the target ASR is then modified to take into account those alternate pronunciations for each phoneme.

The work presented here has been carried out within the scope of the European project \textit{HIWIRE} (\textit{Human Input that Works In Real Environments})
that aims at the development of systems based on ASR to assist human agents in real working conditions.
In \textit{HIWIRE} project, an automated vocal command system is being developed to help aircraft pilots with simple tasks and communications with air traffic control towers (ATCT).
As the communications between ATCT and pilots have to be conducted in English language, the application will be used by non-native speakers.
Thus, the ASR needs to be adjusted to foreign accents of English speech.

We have already presented a new automated approach for non-native speech recognition that uses a ``phonetic confusion'' between SL and NL phones \cite{ES00}.
As non-native speakers tend to pronounce phones in a manner similar to their mother tongue \cite{Ladefoged}, we have used NL acoustic models to represent the non-native pronunciations.
In the work presented here, we test the ``phonetic confusion'' with several other combinations of acoustic model sets.
That is, in the pronunciation modelling, we aim at employing acoustic models that have been acoustically adapted to the foreign accent.
In the next sections, we will describe the ``phonetic confusion'' along with several other foreign accent adaptation techniques that we have used.
Afterwards, we will present and discuss the results of our experiments on the \textit{HIWIRE} database.

\section{Acoustic Adaptation to Foreign Accents}


In the ideal case for automatic foreign accented speech recognition, a large NL-accented SL speech database would be utilized in order to train specific models.
Unfortunately, it would not be feasible to record large enough non-native speech databases for each SL/NL couples.
Nonetheless, relatively small foreign-accented speech corpora are available and could be used efficiently to modify the pretrained SL models.

In the next sections, We describe several approaches of acoustic adaptation to non-native accents.
We aim at adapting pre-trained SL acoustic models on foreign speech in order to capture the non-native accent.
The term ``canonical acoustic models'' refers to the standard models trained on native speech.

\subsection{MLLR and MAP Adaptation to Foreign Accents}
\label{MLLR_MAP_adapt}

Acoustic model adaptation using MLLR (Maximum Likelihood Linear Regression) or MAP (Maximum A Posteriori) methods is widely used for speaker adaptation.

MLLR and MAP techniques have also been employed in non-native accent adaptation for foreign speech recognition \cite{ESTomokiyo}.
We use these techniques to capture non-native accents by adapting canonical SL (English) acoustic models on foreign accented SL speech.
For each NL, the canonical acoustic models of the SL ASR system are adapted in a supervised fashion on non-native SL speech uttered by speakers sharing the same NL origin.
That is, for MLLR adaptation to a foreign accent, we use a one-pass supervised adaptation of the canonical SL models on NL-accented SL speech.
On the other hand, for the MAP method, we chose to performed MLLR followed by MAP supervised adaptation in order to improve the accuracy of the resulting models.
We obtain two sets of speaker independent models adapted (by MLLR and MAP) to NL-accented SL speech. 

\subsection{Model Re-estimation on Foreign Accented Speech}
\label{Model_Re_estimation}

A full training of NL-accented SL acoustic models would not be possible on relatively small non-native speech corpora. 
Nonetheless, those databases could be efficiently employed to re-estimate pre-trained acoustic models in order to capture the non-native accent.
The canonical SL models can be used as starting point in the set-up of NL-accented SL models.
That is, additional Baum-Welch re-estimation steps could be applied on the canonical SL HMMs using those databases.
The canonical SL models are not perfectly fit for foreign-accented SL speech, but they could be a good starting point in the training of NL-accented SL acoustic models.

\section{Pronunciation Modelling}
\label{pronunc_modelling}

Pronunciation modelling of non-native accents consists in identifying the errors that foreign speakers produce and taking those alternate pronunciations into account.
Detection of these errors or deviant pronunciations can be achieved either by an expertise entailing human phoneticians as in \cite{ES1}, or through a data driven approach such as the following.

We presented in \cite{ES00} an automated approach for foreign speech recognition that uses two sets of acoustic models : SL HMM set and NL HMM set.
The first set of models, SL HMM set, consists of the canonical acoustic units that form the canonical pronunciations (i.e. what should have been uttered).
The second set, NL HMM set, is made up of the acoustic units that will be used to represent non-native pronunciations (i.e. what was actually uttered by non-native speakers).
The method we proposed associates to each SL phone several sequences of NL phones. 
First, a phonetic alignment with the SL phones and a phonetic recognition with NL phones are performed on a non-native speech database.
Afterwards, the two transcriptions that resulted from the previous operations are time-aligned in order to associate each SL phone to the sequence of NL phones that occurred in the same time interval.
Finally, only the most probable (frequent) associations are taken into account to form what we call ``phonetic confusion rules''.

\begin{figure}[htpb]
\centerline{\epsfig{figure=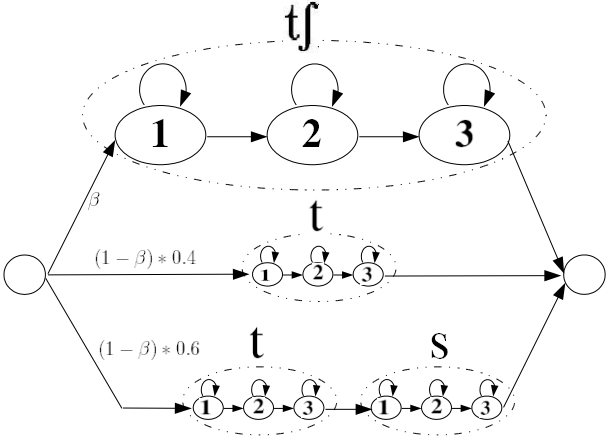,width=80mm}}
\caption{{\it Adding HMM state paths to the model of the English phone \textipa[tS] in the case of Greek accent ($\beta$ is a weight).}}
\label{figure_multi_chemins}
\vspace{-0.3cm}
\end{figure}

The second step of the approach consists in inserting the knowledge acquired in the above procedure into the ASR system.
We chose to modify the HMM of each SL phone $P^s$ by including alternate state paths that represent the deviant pronunciations.
Every new paths corresponds to a confusion rule $R$ related to $P^s$ : it is the concatenation of the HMM models of NL phone sequence associated to $P^s$ in the rule $R$.
This way, the modified HMM model represents the canonical pronunciation of $P^s$ as well as its different alternate forms.
Figure~\ref{figure_multi_chemins} illustrates the modification of the English phone
\begin{minipage}{0.4cm}{\scalebox{0.16}{\includegraphics{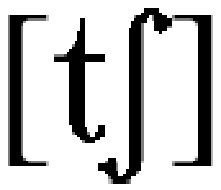}}}\end{minipage}
according to the following extracted confusion rules when modelling Greek accented English :

\begin{tabular}{lll}
					&- \begin{minipage}{0.4cm}{\scalebox{0.16}{\includegraphics{fig_tche.eps}}}\end{minipage} $\rightarrow$ [t]      & $P = 0.4 $\\
					&- \begin{minipage}{0.4cm}{\scalebox{0.16}{\includegraphics{fig_tche.eps}}}\end{minipage} $\rightarrow$ [t] [s]  & $P = 0.6 $\\
\end{tabular}

As non-native speakers tend to pronounce some phones as in their native language \cite{Ladefoged}, we chose to represent the alternate pronunciations of an SL phone as sequences of NL phones.
That is, we chose the SL and NL sets of canonical acoustic units as the first and second sets of models in the pronunciation modelling.
However, other couples of sets of acoustic models could be used in this accent modelling.


In the work presented here, we propose to use different couples of acoustic model sets in the pronunciation modelling in order to enhance the recognition accuracy.
As described above, the pronunciation modelling we developed uses two sets of acoustic models : the first set contains the HMMs in which the canonical pronunciations are expressed, and the second set contains the HMMs in which the alternate pronunciations will be expressed.
Instead of SL and NL HMMs as the first and second set of models, we propose to use SL models that have been acoustically adapted to the foreign accent through MLLR, MAP or re-estimation techniques.
We also propose to use, as the second set of models, NL models that have been acoustically adapted to the non-native accent.
Indeed, those models are better suited to the non-native speech and could achieve a better pronunciation modelling, and thus better recognition results.


\section{Experiments}

\subsection{Experimental conditions}

The work presented here has been carried out within the scope of the European project \textit{HIWIRE} which aims at developing an automated vocal command system designed to help aircraft pilots in their tasks and communications with air traffic control towers (ATCT).
For that matter, a non-native English speech database has been recorded in clean conditions and with 16Khz sampling frequency.
It is composed of 31 French, 20 Greek, 20 Italian, and 10 Spanish speakers each of them uttered 100 sentences.
The grammar is a command language that complies to the communication protocols between ATCT and pilots.
The vocabulary is composed of 134 different words.
We chose 13 MFCC coefficients and their first and second time derivatives as acoustic parameters.
We used 3 state HMM monophones as acoustic models with 128 Gaussian mixtures for all our experiments (except the models described in section \ref{Model_Re_estimation} which had 64 Gaussians).
The English monophones were trained on the \textit{TIMIT} corpus.
The French, Greek, Italian, and Spanish sets of monophones were trained on respective native speech databases.

In our experiments we have used both a constrained grammar and a word-loop grammar.
We adopted the cross-validation approach in our tests in order to virtually increase the size of the database.
In all the tests, the adaptation techniques have been carried out separately for each one of the native languages : French, Greek, Italian and Spanish.
That is, to test an accent adaptation approach on a speaker $X$ having NL as native language, the NL-accented English database (without the utterances of $X$) is used.
All the MLLR and MAP adaptations to the foreign accent that we have performed were done in a supervised fashion.
We also tested offline speaker adaptation through MLLR and MAP supervised techniques.
For MLLR speaker adaptation, we used a global regression class.
For MAP speaker adaptation, we chose to perform an MLLR adaptation prior to MAP.
When applying speaker adaptation, half of the recorded speech of the underlying speaker was used to adapt the models and the rest for the testing.
In all the tests, the factor $\beta$ has been set to $0.5$ (see section \ref{pronunc_modelling}).

\subsection{Acoustic adaptation and pronunciation modelling}

In early experiments, we tested the pronunciation modelling through ``phonetic confusion'' between canonical English and canonical native monophones.
That is, to each canonical SL phone are associated several sequences of canonical NL phones as described in section \ref{pronunc_modelling}.
We also tested the MLLR, MAP and re-estimation techniques to adapt canonical English models to the foreign accent. 

In the following, ``baseline'' denotes the SL-ASR (English acoustic models) system without any modifications.
The ``phonetic confusion'' between canonical English and native models is referred to as ``Confusion3''.
The MLLR and MAP techniques to adapt the canonical English models to the foreign accent (see section \ref{MLLR_MAP_adapt}) are referred to as ``MLLR-ACC'' and ``MAP-ACC'' respectively.
Similarly, the re-estimation approch as described in section \ref{Model_Re_estimation} is denoted as ``Re-estimation''.
Table \ref{table_first_test} summarizes the results of the latter techniques with the constrained and word-loop grammars respectively.
The results are in terms of word and sentence error rates (\textit{WER, SER})
In order to simplify the comparisons, the table contents are sorted by the SER score of the MAP speaker adaptation condition (last column).
Compared to the baseline system, every adaptation method achieved significant improvements.
When no speaker adaptation is applied, the \textit{WER} reduction varies from 19.5\% to 62.5\% (relative) with the constrained grammar and varies from 17.7\% to 55.8\% with the free grammar.
This error reduction is less important when MLLR and MAP speaker adaptations are performed and reaches 42.3\% and 39.5\% with the constrained and word-loop grammars respectively.
Furthermore, when speaker adaptation is applied, the performance of the ``MLLR-ACC'' approach is close to the baseline, which suggests that the MLLR acoustic adaptation is not efficient in the non-native accent modelling.
This meets the results announced by Tomokiyo et al. \cite{ESTomokiyo} and Clarke et al. \cite{ESClark} concerning the relative inefficiency of MLLR with non-native accents.
As can be seen in table \ref{table_first_test}, the pronunciation modelling has achieved significant ameliorations in comparison to the baseline in all testing conditions.
The ``phonetic confusion'' approach is outperformed by the ``MAP-ACC'' and ``Re-estimation'' acoustic adaptation techniques.

\begin{table} [htpb]
\vspace{-0.3cm}
\caption{\label{table_first_test} {\it Results of acoustic modelling and the ``phonetic confusion'' between canonical English and native models.}}
\centerline{
\begin{small}
\begin{tabular}{|l|c|c|c|c|c|c|}
\hline
	& \multicolumn{2}{|c|}{\it
	\begin{minipage}{1.3cm}
	\scriptsize No Speaker \\
	Adaptation
	\end{minipage}
	}
	&
	\multicolumn{2}{|c|}{\it
	\begin{minipage}{1.5cm}
	\scriptsize MLLR Speaker \\
	Adaptation
	\end{minipage}
	}
	&
	\multicolumn{2}{|c|}{\it
	\begin{minipage}{1.5cm}
	\scriptsize MAP Speaker \\
	Adaptation
	\end{minipage}
	}
	\\ \hline
\textit{\normalsize System} & \textit{\scriptsize WER} & \textit{\scriptsize SER} & \textit{\scriptsize WER} & \textit{\scriptsize SER} & \textit{\scriptsize WER} & \textit{\scriptsize SER}
\\
\hline
\multicolumn{7}{l}{Constrained grammar :} \\
\hline
{\small Baseline     }& 7.2 &14.6 & 4.8 &10.6 & 2.6 & 6.3 \\ \hline
{\small MLLR-ACC     }& 5.8 &12.1 & 4.6 & 9.9 & 2.8 & 6.2 \\ \hline
{\small Confusion3   }& 4.8 &10.9 & 3.1 & 7.5 & 2.0 & 5.0 \\ \hline
{\small Re-estimation}& 3.0 & 6.4 & 2.3 & 5.1 & 1.9 & 4.1 \\ \hline
{\small MAP-ACC      }& 2.7 & 6.2 & 2.0 & 4.8 & 1.5 & 3.7 \\ \hline
\multicolumn{7}{l}{Word-loop grammar :} \\
\hline
{\small Baseline     }& 38.5 & 49.9 & 29.1 & 41.4 & 19.7 & 31.7 \\ \hline
{\small MLLR-ACC     }& 32.1 & 43.9 & 27.5 & 40.2 & 19.3 & 31.2 \\ \hline
{\small Confusion3   }& 26.8 & 43.0 & 21.4 & 36.0 & 15.7 & 27.8 \\ \hline
{\small MAP-ACC      }& 18.2 & 30.7 & 16.1 & 27.8 & 12.9 & 23.0 \\ \hline
{\small Re-estimation}& 17.0 & 27.0 & 14.2 & 24.1 & 11.9 & 21.0 \\ \hline
\end{tabular}
\end{small}
}
\vspace{-0.4cm}
\end{table}

\subsection{Combined acoustic and pronunciation modelling}

The next part of our work consists in combining the acoustic adaptation and the pronunciation modelling.
That is, as input model sets in the pronunciation modelling, we use HMM models that have been acoustically adapted to the foreign accent as described in section \ref{pronunc_modelling}.
Table~\ref{table_machin_chose} lists the combinations of acoustic model sets we have used in the accent modelling.
The term ``Native + MLLR'' (resp. ``Native + MAP'') refers to the NL acoustic models that have been acoustically adapted to the foreign accent using non-supervised MLLR (resp. MAP).
That is, a phonetic recognition is performed on NL-accented English speech database using the NL models.
Then, NL models are adapted, on that database, through MLLR and MAP according to the results of the latter recognition.
\begin{table}[htpb]
\caption{\label{table_machin_chose} {\it List of couples of HMM sets used in the pronunciation modelling.}}
\begin{tabular}{|l|l|l|}
\hline
System & First set of models & Second set of models 			\\ \hline\hline
Confusion1 & ``\textit{MLLR-ACC}'' 	& ``\textit{MLLR-ACC}'' 	\\ \hline
Confusion2 & Canonical English 		& Canonical English  		\\ \hline
Confusion3 & Canonical English 		& Canonical Native  		\\ \hline
Confusion4 & ``\textit{MLLR-ACC}'' 	& ``Native + MLLR'' 		\\ \hline
Confusion5 & ``\textit{MAP-ACC}'' 	& ``Native + MAP'' 		\\ \hline
Confusion6 & ``\textit{MAP-ACC}'' 	& ``\textit{MAP-ACC}'' 		\\ \hline
Confusion7 & Canonical English 		& ``Re-estimation''\\ 		\hline
\end{tabular}
\vspace{-0.65cm}
\end{table}

Table \ref{table1} summarizes the results of the pronunciation modelling using the latter couples of HMM models sets.
With both free and constrained grammars, we observe improvements for all the systems compared to the baseline.
Nonetheless, an exception arises to the latter concerning the ``Confusion1'' and ``Confusion2'' approaches which perform worse than the baseline with the MAP speaker adaptation.
This behavior could be explained by the fact that the ``Confusion2'' (resp. ``Confusion1'') pronunciation modelling entails a confusion between identical canonical English models (rep. English models adapted with MLLR to the foreign accent).

Indeed, the results shown in tables \ref{table_first_test} and \ref{table1} support the conclusion that the lack of variability in the models used for the pronunciation modelling penalizes the quality of the resulting models.
That is, the ``phonetic confusion'' between English and native models (both canonical or acoustically adapted to the accent) is more beneficial than a confusion between English models only.

Nonetheless, for MAP acoustic adaptation to the accent, the approach ``Confusion6'' entailing a ``phonetic confusion'' between English models outperforms the approach ``Confusion5'' which consists in a ``phonetic confusion'' between English and native models.
This might be due to the fact that English models used in ``Confusion6'' were adapted in a supervised manner while the native models used in ``Confusion5'' were adapted in a non-supervised fashion.

Another interesting result is the performance of the re-estimated English models described in section \ref{Model_Re_estimation}.
With the constrained grammar, these models lead to significant improvements while they achieved the best results with the free grammar.
Moreover, the pronunciation modelling in ``Confusion7'' approach allowed further improvements and lead to the best results in all conditions.
This suggests that the re-estimation approach on a small adaptation corpora allows a good modelling of the non-native accent.

\begin{table} [t]
\caption{\label{table1} {\it Results of pronunciation modelling using acoustic models adapted to the foreign accent.}}
\centerline{
\begin{small}
\begin{tabular}{|l|c|c|c|c|c|c|}
\hline
	& \multicolumn{2}{|c|}{\it
	\begin{minipage}{1.3cm}
	\scriptsize No Speaker \\
	Adaptation
	\end{minipage}
	}
	&
	\multicolumn{2}{|c|}{\it
	\begin{minipage}{1.5cm}
	\scriptsize MLLR Speaker \\
	Adaptation
	\end{minipage}
	}
	&
	\multicolumn{2}{|c|}{\it
	\begin{minipage}{1.5cm}
	\scriptsize MAP Speaker \\
	Adaptation
	\end{minipage}
	}
	\\ \hline
\textit{\normalsize System} & \textit{\scriptsize WER} & \textit{\scriptsize SER} & \textit{\scriptsize WER} & \textit{\scriptsize SER} & \textit{\scriptsize WER} & \textit{\scriptsize SER}
\\
\hline
\multicolumn{7}{l}{Constrained grammar :} \\
\hline
Confusion1	& 5.3 &10.2 & 4.4 & 9.5 & 3.1 & 7.2 \\ \hline
Confusion2	& 5.8 &11.8 & 4.4 & 9.4 & 2.9 & 6.5 \\ \hline
Baseline	& 7.2 &14.6 & 4.8 &10.6 & 2.6 & 6.3 \\ \hline
Confusion3	& 4.8 &10.9 & 3.1 & 7.5 & 2.0 & 5.0 \\ \hline
Confusion4	& 3.5 & 8.1 & 3.1 & 5.2 & 2.1 & 4.8 \\ \hline
Confusion5	& 2.8 & 6.4 & 2.3 & 5.2 & 1.8 & 4.1 \\ \hline
Confusion6	& 2.8 & 6.5 & 2.2 & 5.0 & 1.7 & 4.1 \\ \hline
Confusion7	& 2.1 & 5.0 & 1.6 & 4.0 & 1.4 & 3.2 \\ \hline
\multicolumn{7}{l}{Word-loop grammar :} \\
\hline
Baseline	& 38.5 & 49.9 & 29.1 & 41.4 & 19.7 & 31.7 \\ \hline
Confusion1	& 27.1 & 41.6 & 23.5 & 37.5 & 18.1 & 30.5 \\ \hline
Confusion2	& 29.0 & 42.9 & 23.3 & 37.1 & 17.4 & 29.6 \\ \hline
Confusion3	& 26.8 & 43.0 & 21.4 & 36.0 & 15.7 & 27.8 \\ \hline
Confusion4	& 22.9 & 37.3 & 20.1 & 34.0 & 15.1 & 26.8 \\ \hline
Confusion5	& 17.3 & 30.5 & 15.6 & 28.1 & 13.2 & 24.1 \\ \hline
Confusion6	& 17.3 & 29.7 & 15.6 & 27.1 & 12.9 & 23.2 \\ \hline
Confusion7	& 15.5 & 26.0 & 13.5 & 23.8 & 11.4 & 20.8 \\ \hline
\end{tabular}
\end{small}
}
\vspace{-0.5cm}
\end{table}

\vspace{-0.2cm}
\section{Conclusion}

In this paper, we presented several non-native accent approaches based on the combination of pronunciation modelling and acoustic adaptation.
We used MLLR, MAP, and model re-estimation to acoustically adapt the English models to the non-native accent.
The pronunciation modelling we have developed consists in associating several sequences of native language phones to each spoken language phone.
We have also combined the acoustic adaptation and the pronunciation modelling by using acoustically adapted HMMs in the accent modelling.
The obtained results suggest that MLLR adaptation to the non-native accent is relatively inefficient.
Moreover, our experiments show that using both spoken and native language models leads to more accurate modelling of foreign accents.
Finally, we can note that model re-estimation technique combined with pronunciation modelling achieved the best results.

\vspace{-0.2cm}
\section{Acknowledgments}

This work was partially funded by the European project \textit{HIWIRE} (\emph{Human Input that Works In Real Environments}), contract number 507943,
\emph{sixth framework program, information society technologies}.

\vspace{-0.2cm}

\end{document}